\documentclass[twoside]{article}
\usepackage[preprint]{aistats2022}

\usepackage[round]{natbib}

\usepackage[utf8]{inputenc} 
\usepackage[T1]{fontenc}    
\usepackage{url}            
\usepackage{booktabs}       
\usepackage{amsfonts}       
\usepackage{nicefrac}       
\usepackage{microtype}      
\usepackage{amsthm}
\usepackage{wasysym}
\usepackage{url}
\usepackage{wrapfig}
\usepackage{xr-hyper}

\usepackage{hyperref}       
\hypersetup{
  colorlinks=true,
  citecolor=black,
  linkcolor=black,
  urlcolor=black,
  filecolor=black,
}

\usepackage{amsmath}
\usepackage{amssymb}

\usepackage{lipsum}

\usepackage[capitalise]{cleveref}
\crefname{appendix}{Supplement}{Supplements}

\usepackage{subcaption}
\usepackage[dvipsnames]{xcolor}
\definecolor{notecolor}{RGB}{0,82,147}

\definecolor{blu}{RGB}{50,106,170}
\definecolor{gre}{RGB}{63,129,71}
\definecolor{drd}{RGB}{179,45,38}
\definecolor{pur}{RGB}{81,64,130}

\definecolor{c1}{HTML}{107D79}
\definecolor{c2}{HTML}{FF9933}
\definecolor{c3}{HTML}{1F77B4}
\definecolor{c4}{HTML}{D62728}
\definecolor{c5}{HTML}{9467BD}
\definecolor{c6}{HTML}{8C564B}
\definecolor{c7}{HTML}{E377C2}
\definecolor{c8}{HTML}{7F7F7F}
\definecolor{c9}{HTML}{BCBD22}
\definecolor{c10}{HTML}{17BECF}

\definecolor{gruvboxyellow}{HTML}{D79921}
\definecolor{gruvboxblue}{HTML}{458588}

\definecolor{cred}{HTML}{a54242}
\definecolor{cgreen}{HTML}{8c9440}
\definecolor{cyellow}{HTML}{de935f}
\definecolor{cblue}{HTML}{5f819d}

\newcommand{\burden}[1]{}  
\newtheorem{remark}{Remark}

\theoremstyle{definition}

\newcommand{\T}{^{\intercal}}
\renewcommand{\d}{\:\mathrm{d}}

\newcommand{\Z}{\mathcal{Z}}
\newcommand{\I}{\mathcal{I}}

\newcommand{\IWP}{\text{IWP}}

\renewcommand{\tilde}{\widetilde}

\usepackage{tikz}
\usetikzlibrary{bayesnet}
\usetikzlibrary{positioning}

\begin{document}

\runningtitle{Pick-and-Mix Information Operators for Probabilistic ODE Solvers}
\runningauthor{Nathanael Bosch, Filip Tronarp, Philipp Hennig}

\twocolumn[

\aistatstitle{Pick-and-Mix Information Operators\\ for Probabilistic ODE Solvers}

\aistatsauthor{ Nathanael Bosch\textsuperscript{1} \And Filip Tronarp\textsuperscript{1} \And Philipp Hennig\textsuperscript{1,2}}

\aistatsaddress{
  \textsuperscript{1}University of Tübingen\\
  \textsuperscript{2}Max Planck Institute for Intelligent Systems, Tübingen, Germany
}
]

\begin{abstract}
  Probabilistic numerical solvers for ordinary differential equations compute posterior distributions over the solution of an initial value problem via Bayesian inference.
  In this paper, we leverage their probabilistic formulation to seamlessly include additional information as general likelihood terms.
  We show that second-order differential equations should be directly provided to the solver, instead of transforming the problem to first order.
  Additionally, by including higher-order information or physical conservation laws in the model, solutions become more accurate and more physically meaningful.
  Lastly, we demonstrate the utility of flexible information operators by solving differential-algebraic equations.
  In conclusion, the probabilistic formulation of numerical solvers offers a flexible way to incorporate various types of information, thus improving the resulting solutions.
\end{abstract}

\section{INTRODUCTION}
\label{sec:introduction}

\begin{figure}[ht!]
  \centering
  \includegraphics[width=1.0\linewidth]{%
    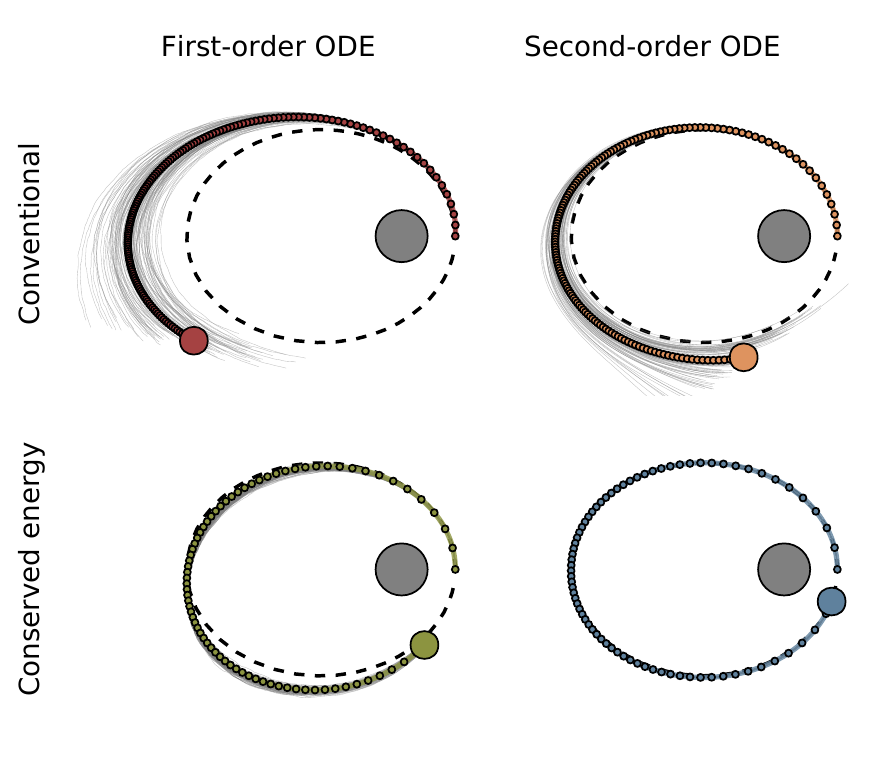}
  \caption{
    \emph{Faithful modeling of ODE information improves probabilistic solutions.}
    Informing the solver about the second-order structure of the Kepler problem (\textcolor{cyellow}{\newmoon}) increases the accuracy over its first-order counterpart (\textcolor{cred}{\newmoon}).
    Adding physical information about the conservation of energy and angular momentum greatly improves the solution, \emph{even for increased step sizes} (\textcolor{cgreen}{\newmoon}).
    Full information leads to the best results (\textcolor{cblue}{\newmoon}).
  }
  \label{fig:1}
\end{figure}

Throughout science and engineering, dynamical systems are frequently described with ordinary differential equations (ODEs).
But in many systems of interest, the differential equation harbors additional information not directly accessible to the numerical algorithm used to solve it.
For example, physical systems often follow high-order dynamics and preserve quantities such as energy, mass, or angular momentum.
To efficiently compute meaningful solutions, practitioners have to carefully choose from a wide range of numerical solvers, such as
Runge--Kutta methods \citep{hairer2008solving},
Nyström methods for second-order ODEs \citep{nystrom1925numerische},
structure-preserving integrators \citep{hairer2006geometric},
and many more.
Each of these methods can be seen as a laboriously custom-designed way to encode specific kinds of information.
In this paper, we present a more flexible, unified approach to include additional knowledge into numerical ODE solutions, by leveraging the framework of \emph{probabilistic numerics}.

In probabilistic numerics
\citep{hennig15_probab_numer_uncer_comput, DBLP:journals/sac/OatesS19},
numerical problems are formulated as problems of probabilistic inference.
Probabilistic numerical methods return \emph{distributions over solutions}.
Such methods can quantify their own approximation error through samples and other structured quantities -- a functionality typically not provided by classic numerical methods.
This paper builds on probabilistic numerical ODE solvers based on Bayesian filtering and smoothing
\citep{schober16_probab_model_numer_solut_initial_value_probl,tronarp18_probab_solut_to_ordin_differ}.
These ``ODE filters'' have been shown to converge with polynomial rates
\citep{kersting18_conver_rates_gauss_ode_filter,tronarp20_bayes_ode_solver}
and their efficiency has been demonstrated on
a range of both non-stiff and stiff problems
\citep{kraemer20_stabl_implem_probab_ode_solver,bosch20_calib_adapt_probab_ode_solver}.

\paragraph{Contributions}
Probabilistic ODE solvers are defined by two parts: the prior and the likelihood.
In the basic version of such solvers, the likelihood is completely defined by the vector field.
But, as we show in this work, their formulation is sufficiently flexible to allow for a much richer language.
By formulating the likelihood in terms of flexible information operators, information about higher-order derivatives and conserved quantities can be represented with the same semantics as the ODE information itself.
We demonstrate the utility of the proposed framework in four case studies:
\begin{enumerate}
  \item
    \emph{Second-order differential equations:}
    Solving second-order ODEs directly, instead of transforming them to first order, greatly improves the efficiency of probabilistic solvers.
  \item
    \emph{Additional second-derivative information:}
    Information about higher-order derivatives can be additionally included in the joint inference process to increase the solution accuracy.
  \item
    \emph{Systems with conserved quantities:}
    By including conservation laws into the model, probabilistic solutions become not only more accurate but also more physically meaningful.
  \item
    \emph{Differential-algebraic equations (DAEs):}
    With the corresponding information operator, probabilistic solvers can be extended to DAEs.
\end{enumerate}

\section{PROBABILISTIC ODE SOLVERS}
\label{sec:background}
This section introduces filtering-based probabilistic ODE solvers.
Consider an initial value problem (IVP)
\begin{equation} \label{eq:ode1}
  \dot{y} = f(y(t), t), \qquad \forall t \in [0, T],
\end{equation}
with vector field
\(f: \mathbb{R}^{d+1} \to \mathbb{R}^d\)
and initial value
\(y(0) = y_0 \in \mathbb{R}^d\).
Instead of computing a single point estimate (as done by classic numerical algorithms), ODE filters compute \emph{probabilistic} ODE solutions.
That is, they approximate posterior distributions of the form
\begin{equation} \label{eq:probabilistic-ode-solution}
  p \left( y(t) ~|~ y(t_0) = y_0, \{\dot{y}(t_n) = f(y(t_n), t_n)\}_{n=0}^N \right),
\end{equation}
for a chosen time-discretization \(\{t_n\}_{n=1}^N\).
Thereby, they estimate not only the ODE solution, but also the unavoidable, global approximation error that arises due to discretization.

In the following, we pose the probabilistic numerical ODE solution as a problem of Bayesian state estimation, the solution of which can be efficiently approximated with extended Kalman filtering.
For a more thorough introduction we refer to \citet{tronarp18_probab_solut_to_ordin_differ}.

\subsection{Numerical ODE Solutions As Inference}
\label{sec:model}
\paragraph{Integrated Wiener Process Priors}
\label{sec:prior}
\emph{A priori}, we model the unknown ODE solution \(y(t)\) by a \(q\)-times integrated Wiener process (IWP).
More precisely, define
\begin{equation}
  Y(t) = \left[Y^{(0)}(t), Y^{(1)}(t), \dots, Y^{(q)}(t)\right]
\end{equation}
as the solution of a linear, time-invariant stochastic differential equation of the form
\begin{subequations} \label{eq:continuous-prior}
  \begin{align}
  \d Y^{(i)}(t) &= Y^{(i+1)}(t) \d t,  \qquad i = 0, \dots q-1, \\
  \d Y^{(q)}(t) &= \Gamma^{1/2} \d W(t), \\
    Y(0) &\sim \mathcal{N} \left( \mu_0, \Sigma_0 \right),
  \end{align}
\end{subequations}
driven by a \(d\)-dimensional Wiener process \(W\).
The matrix \(\Gamma^{1/2}\) is the symmetric square-root of some positive semi-definite matrix \(\Gamma \in \mathbb{R}^{d \times d}\) and
\(\mu_0 \in \mathbb{R}^{d(q+1)}\),
\(\Sigma_0 \in \mathbb{R}^{d(q+1) \times d(q+1)}\) are the initial mean and covariance.
Then, \(Y^{(i)}\) models the \(i\)-th derivative of unknown ODE solution \(y\) and we write \(y \sim \IWP(q)\).

\paragraph{Discrete-Time Transitions}
The process \(Y(t)\) satisfies transition densities
\citep{sarkka_solin_2019}
\begin{equation} \label{eq:discrete-prior}
  Y(t + h) \,|\, Y(t) \sim \mathcal{N} \left( A(h) Y(t), Q(h) \right).
\end{equation}
The matrices \(A(h), Q(h) \in \mathbb{R}^{d(q+1) \times d(q+1)}\) denote the transition matrix and the process noise covariance.
For the chosen \(\IWP(q)\) prior, it holds
\begin{equation}
    A(h) = \breve{A}(h) \otimes I_d, \qquad Q(h) = \breve{Q}(h) \otimes \Gamma,
\end{equation}
and the matrices \(\breve{A}(h), \breve{Q}(h) \in \mathbb{R}^{(q+1) \times (q+1)}\) are known in closed form
\citep{kersting18_conver_rates_gauss_ode_filter}:
\begin{subequations}
  \begin{align}
    \breve{A}_{ij}(h) &= \mathbb{I}_{i \leq j} \frac{h^{j-1}}{(j-i)!}, \\
    \breve{Q}_{ij}(h) &= \frac{h^{2q+1-i-j}}{(2q+1-i-j)(q-i)!(q-j)!}.
  \end{align}
\end{subequations}

\paragraph{Measurement Process}
To relate the prior process to the ODE solution,
we define a measurement model in terms of an \emph{information operator}
\citep{cockayne17_bayes_probab_numer_method,tronarp18_probab_solut_to_ordin_differ},
similar to the likelihood models used in gradient matching  \citep{calderhead09_acc_bayes_inf,wenk19_odin}.
Define
\begin{equation}\label{eq:Z1}
  \Z[y](t) := \dot{y}(t) - f(y(t), t).
\end{equation}
The operator \(\Z\) maps the true ODE solution (see \cref{eq:ode1}) to a known quantity, namely the zero function; \(\Z[y] \equiv 0\).
On the other hand, the action of the information operator on the process \(Y\) can be expressed in terms of the following non-linear function
\begin{equation}\label{eq:z1}
  z \left( t, Y \right) := \Z[Y^{(0)}](t) = Y^{(1)}(t) - f \left( Y^{(0)}(t), t \right).
\end{equation}
Once again, if \(Y^{(0)}\) solves the ODE (\cref{eq:ode1}) exactly, we have \(z(t, Y) \equiv 0\).
Consequently, \emph{inferring} the true ODE solution \(y\) reduces to conditioning the prior \(Y\) on
\(z(t, Y) = 0\).
This inference problem is the subject of the next section.

\subsection{Approximate Gaussian Inference}
\label{sec:inference}
To enable tractable inference, we discretize time to a grid \(\{t_n\}_{n=1}^N \subset [0, T]\) and condition the process \(Y(t)\) only on discrete observations
\(z_n := z(t_n, Y(t_n))=0\).
The resulting non-linear Gauss--Markov regression problem is well-known in the Bayesian filtering and smoothing literature
\citep{sarkka_bayesianfilteringandsmoothing}.
Its solution can be efficiently approximated with the extended Kalman filter (EKF), as Gaussian distributions
\begin{equation}
  p \left( Y(t_n) \,|\, z_{1:n} \right)
  \approx \mathcal{N} \left( \mu_n, \Sigma_n \right).
\end{equation}

In a nutshell, the EKF algorithm proceeds by iterating the following steps
\citep[Section 5.2]{sarkka_bayesianfilteringandsmoothing}:
\begin{itemize}
  \item \texttt{PREDICT:}
    Given
    \(Y(t_{n}) ~|~ z_{1:n} \sim \mathcal{N} \left( \mu_{n}, \Sigma_{n} \right)\)
    and the Gaussian transitions of \cref{eq:discrete-prior}, we can extrapolate to
    \(Y(t_{n+1}) ~|~ z_{1:n} \sim \mathcal{N} \left( \mu_{n+1}, \Sigma_{n+1} \right)\),
    with%
    \begin{subequations} \label{eq:predict}
      \begin{align}
        \mu_{n+1}^- &= A(h_{n}) \mu_{n}, \\
        \Sigma_{n+1}^- &= A(h_{n}) \Sigma_{n} A(h_{n})\T + Q(h_{n}),
      \end{align}
    \end{subequations}
    where \(h_n := t_{n+1} - t_n\).
  \item \texttt{UPDATE:}
    To include information about the new measurement \(z_{n+1}\) into \(Y(t_{n+1})\) approximate
    \( Y(t_{n+1}) ~|~ z_{1:n+1} \sim \mathcal{N} \left( \mu_{n+1}, \Sigma_{n+1} \right) \), with
    \begin{subequations} \label{eq:update}
      \begin{align}
        \hat{z}_{n+1} &= z \left( t_{n+1}, \mu_{n+1}^- \right) \\
        S_{n+1} &= H_{n+1} \Sigma_{n+1}^- H_{n+1}\T, \\
        K_{n+1} &= \Sigma_{n+1}^- H_{n+1}\T S_{n+1}^{-1}, \\
        \mu_{n+1} &= \mu_{n+1}^- + K_{n+1} (z_{n+1} - \hat{z}_{n+1}), \\
        \Sigma_{n+1} &= \Sigma_{n+1}^- - K_{n+1} S_{n+1} K_{n+1}\T.
      \end{align}
    \end{subequations}
    In a standard EKF, the matrix \(H_{n+1}\) denotes the Jacobian of the measurement model \(z\), evaluated at \(\mu_{n+1}^-\).
    For \(z\) as defined in \cref{eq:z1}, we have
    \( H_n := E_1 - J_f(E_0 \mu_n, t_n) E_0 \),
    where the matrices \(E_i \in \mathbb{R}^{d \times d(q+1)}\)
    denote projection matrices to the \(i\)-th component of the state \(Y\), that is \(E_i Y = Y^{(i)}\).
    We call the resulting ODE solver \texttt{EK1}
    \citep{tronarp18_probab_solut_to_ordin_differ}.
    Alternatively,
    \citet{schober16_probab_model_numer_solut_initial_value_probl} use
    a zeroth-order approximation of the vector field, i.e.~\( H_n := E_1 \).
    We refer to this solver as \texttt{EK0}.
\end{itemize}

\begin{remark}[Smoothing]
  A Rauch--Tung--Striebel backward pass turns the filtering distribution into a smoothing posterior
  \citep{sarkka_bayesianfilteringandsmoothing}.
  %
  %
  %
  At the final time point, the filtering and smoothing posteriors coincide.
\end{remark}

\begin{remark}[Alternative inference schemes]
  The unscented Kalman filter \citep{1271397} can be used for Gaussian filtering, but requires multiple evaluations of the vector field at each time step.
  Particle filtering
  \citep{sarkka_bayesianfilteringandsmoothing}
  can provide more descriptive, non-Gaussian ODE posterior estimates
  \citep{tronarp18_probab_solut_to_ordin_differ}.
  But, similarly to sampling-based approaches to probabilistic ODE solutions
  \citep{chkrebtii13_bayes_solut_uncer_quant_differ_equat,Conrad2017,Abdulle2020,NEURIPS2018_228b2558},
  this expressivity comes at increased computational cost.
  In comparison, the EKF provides computationally efficient approximate inference.
\end{remark}

\subsection{Practical Considerations}
\label{sec:practical-considerations}
\begin{table*}[t]
  \caption{\label{table:operators}Common problem settings and corresponding information operators.}
  \centering
  \begin{tabular}{rll}
    \toprule
    Description & Equation & Information operator\\
    \midrule
    First-order ODE & \(\dot{y}(t) = f \left( y(t), t \right) \) & \(z(t, Y) := Y^{(1)} - f \left(Y^{(0)}, t \right)\)\\
    Second-order ODE & \(\ddot{y}(t) = f \left( \dot{y}(t), y(t), t \right) \) & \(z(t, Y) := Y^{(2)} - f \left( Y^{(1)}, Y^{(0)}, t \right) \)\\
    Mass matrix DAE & \(M\dot{y}(t) = f \left( y(t), t \right) \) & \(z(t, Y) := MY^{(1)} - f \left(Y^{(0)}, t \right)\)\\
    \midrule
    Invariances & \(g \left( y(t), \dot{y}(t) \right) = 0\) & \(z(t, Y) := g \left( Y^{(0)}, Y^{(1)} \right) \)\\
    Chain rule & \(\ddot{y}(t) = J_f \left( y(t) \right) \cdot \dot{y}(t)\) & \(z(t, Y) := Y^{(2)} - J_f \left( Y^{(0)} \right) \cdot Y^{(1)}\)\\
    \bottomrule
  \end{tabular}
\end{table*}
\paragraph{Calibration}
The posterior covariances returned by the solver depend on the choice of diffusion parameter \(\Gamma\)
(recall \cref{eq:continuous-prior}).
Good uncertainty quantification therefore requires the estimation of \(\Gamma\).
In ODE filters, this is usually done by approximately maximizing the marginal likelihood of the observed data \(p(z_{1:N})\)
\citep{tronarp18_probab_solut_to_ordin_differ}.
This procedure also extends to more general, time-varying diffusion models \(\Gamma_n\) which have been proposed for greater flexibility
(and for step-size adaptation; see below)
\citep{schober16_probab_model_numer_solut_initial_value_probl}.
Refer to
\citet{bosch20_calib_adapt_probab_ode_solver} for more detail.
%

\paragraph{Step-Size Adaptation}
In practice, computationally efficient ODE solvers typically rely on adaptive step-size selection
\citep[Chapter II.4]{hairer2008solving}.
We follow the presentation of \citet{bosch20_calib_adapt_probab_ode_solver} and control a local error estimate, derived from the measurement \(z\), with a PI control algorithm \citep{Gustafsson1988}.


\section{INFORMATION OPERATORS}
\label{sec:information-operators}
The previous section established ODE filters as efficient algorithms for computing probabilistic numerical solutions of first-order ODEs.
In the following, we extend their formulation to a broader class of problems and include additional types of information, by generalizing the underlying information operators.


The vector-field information enters the inference problem through the specified measurement model:
\(f\) (recall \cref{eq:ode1}) only appears in the information operator \(\Z\) (respectively \(z\); see \cref{eq:Z1,eq:z1}).
%
However, the approximate inference algorithm itself (the EKF; see \cref{sec:inference}) does not rely on the specific form of the measurements;
except for calibration and step-size adaptation, which we separately discuss below.
%
To extend the ODE filter framework, we consider more general information operators, of the form
\begin{equation}
  \Z \in \I_y := \left\{ \Z : \Z [y] \equiv 0 \right\}.
\end{equation}
As before, they map some unknown function of interest \(y\) to the known zero function.
But, this general form is not restricted to first-order ODEs.
For example, given an energy-preserving system with second-order dynamics, we can formulate a corresponding operator to define its probabilistic solution
(as will be shown in \cref{sec:manifold}).
\Cref{table:operators} provides a summary of the problem settings and the corresponding operators considered in this paper,
written in the functional form \(z(t, Y) := \Z[Y^{(0)}](t)\).
Before moving to our case studies, where each model will be explained in more detail, we discuss practical details and implementation.

\paragraph{Inference with Multiple Information Operators}
Some problems of interest provide multiple types of information about the true solution, for example as
additional derivatives (\cref{sec:chainrule}) or physical conservation laws (\cref{sec:manifold}).
Formally, this amounts to an information operator \(\Z \in \I_y\) that can be partitioned as
\({\Z[y] = \left[ \Z_1[y]\T, \Z_2[y]\T \right]\T}\),
with \(\Z_1, \Z_2 \in \I_y\),
and corresponding functional representation
\begin{equation}
  z(t, Y) = \left[ z_1(t, Y)\T,\  z_2(t, Y)\T \right]\T.
\end{equation}
It is still possible to update jointly on both measurement models in a single EKF update step on \(z\); this strategy is chosen in
\cref{sec:chainrule}.
However, performing two separate update steps can sometimes be preferable
\citep{8861457,partitionedupdateKF,RAITOHARJU2017289}.
In this case, each measurement model is linearized separately in the partially updated state.
This strategy is chosen in \cref{sec:manifold}.

\paragraph{Calibration and Step-Size Adaptation}
The approaches for calibration and adaptive step-size selection discussed in \cref{sec:practical-considerations} do not strictly depend on the specific information operator, but they were developed in the context of first-order ODEs
\citep{bosch20_calib_adapt_probab_ode_solver}.
There, the information operator is \(d\)-dimensional, i.e.~\(z(t, Y) \in \mathbb{R}^d\), and describes the \emph{local defect}.
We found that this formulation can be extended to settings with a different problem structure (in this work, second-order ODEs and DAEs), but for settings with multiple sources of information (here, additional derivatives or invariances) special care has to be taken.
To conveniently consider user-specified relative tolerance levels,
the local error should be of the same dimension as the ODE solution.
Thus, in \cref{sec:chainrule,sec:manifold}, only the part of the measurement model that relates to the given differential equation is considered for calibration and step-size adaptation.



\section{CASE STUDIES}
\label{sec:case-studies}

\begin{figure*}[t!]
  \centering
  \includegraphics[width=\textwidth]{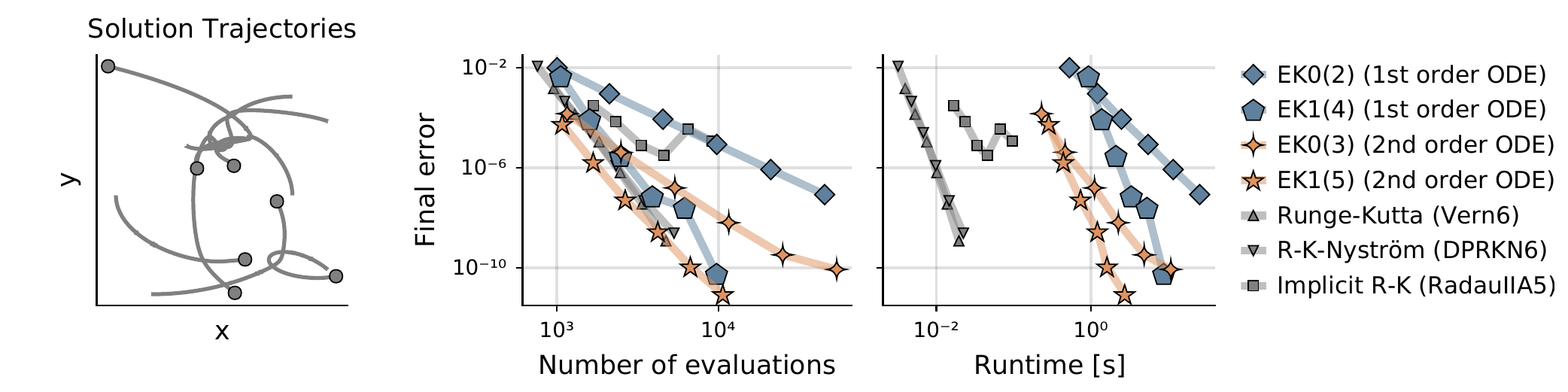}
  \caption{
    \emph{Second-order ODEs should be solved directly.}
    The Pleiades system describes the motion of seven stars in a plane (left).
    Solving this problem directly in second order, compared to solving the equivalent first-order ODE, improves accuracy and efficiency, both in the number of function evaluations (center) and runtime (right).
  }
  \label{fig:ode2}
\end{figure*}

We evaluate the presented framework in four case studies.
First, we apply the probabilistic solver to second-order ODEs.
We investigate the difference between solving such problems directly, by selecting the correct information operator, and solving the algebraically (but not numerically) equivalent first-order ODEs.
Second, we augment the probabilistic numerical solver for first-order ODEs with second-derivative information, which can be computed from the ODE via the chain rule.
Third, we consider Hamiltonian systems in which the total energy is conserved over time, and we evaluate the influence of this information on the probabilistic numerical solution.
Fourth, we demonstrate how probabilistic solvers can be extended to solve semi-explicit differential-algebraic equations.

\paragraph{Implementation}
\label{sec:implementation}
The implementation follows the practices suggested by
\citet{kraemer20_stabl_implem_probab_ode_solver}
and includes exact initialization, preconditioned state transitions, and a square-root implementation.
All experiments are implemented in the Julia programming language
\citep{bezanson2017julia}.
Reference solutions are computed with DifferentialEquations.jl
\citep{rackauckas2017differentialequations}.
All experiments run on a single, consumer-level CPU.
Code for the implementation and experiments is publicly available on GitHub.%
\footnote{Code will be published upon acceptance.}

\subsection{Second-Order Differential Equations}

\label{sec:ode2}
This first case study demonstrates how information about the problem structure, such as the order of the ODE, can improve probabilistic solutions.
To this end, consider
an autonomous, \emph{second}-order ODE
\begin{equation}
  \label{eq:ivp2}
  \ddot{y}(t) = f \left( \dot{y}(t), y(t)\right), \qquad \forall t \in [0, T],
\end{equation}
with vector field
\(f: \mathbb{R}^d \times \mathbb{R}^d \to \mathbb{R}^d\)
and initial values
\(y(0) = y_0\),
\(\dot{y}(0) = \dot{y}_0\).

Second-order ODEs can be transformed to first order by defining a new variable \(\tilde{y} := (\dot{y},y)\).
They can therefore, in principle, be solved by any generic solver.
However, doubling the dimension of the ODE can increase both the solver runtime and memory cost.
Specialized non-probabilistic solvers such as Nyström methods have been specifically developed to circumvent this issue \citep{nystrom1925numerische,hairer2008solving}.
In this section, we follow a similar (but much simpler) approach and present a direct application of probabilistic solvers to second-order ODEs.

The motivation is twofold.
First, a duplication of the ODE dimension leads to a 4x increase in memory cost and 8x runtime, since the EKF algorithm relies on matrix-matrix operations on the state covariances.
Second, the structure of the transformed problem is not a good fit for the integrated Wiener process prior.
After transformation, the first derivative \(\dot{y}\) appears both in \(\tilde{y}\) and \(\frac{\d \tilde{y}}{\d t}\).
It is therefore modeled with both an \(\text{IWP}(q)\) and \(\text{IWP}(q-1)\) prior \emph{at the same time} (recall \cref{sec:prior}).
Both of these shortcomings can be circumvented by solving the second-order problem directly.

\paragraph{Solver Setup}
The second-order ODE (\cref{eq:ivp2}) induces an information operator of the form
\begin{equation}
  z(t, Y) = Y^{(2)} - f \left( Y^{(1)}, Y^{(0)} \right).
\end{equation}
We consider two linearizations:
\begin{equation}
  H := \begin{cases}
    E_2, \qquad&\text{(\texttt{EK0})} \\
    E_2 - \frac{\partial f}{\partial y} \cdot E_0 - \frac{\partial f}{\partial \dot{y}} \cdot E_1, \qquad&\text{(\texttt{EK1})}
  \end{cases}
\end{equation}
named in correspondence to the existing probabilistic solvers for first-order problems presented in \cref{sec:inference}.

\paragraph{Experiment Setup}
We evaluate the solvers on the Pleiades problem
\citep[Chapter II.10]{hairer2008solving},
a system of 14 second-order ODEs that describes the motion of seven stars in a plane
(full problem definition in \cref{sec:prob:pleiades}).
All solvers use adaptive steps and a time-varying diffusion model
\citep{bosch20_calib_adapt_probab_ode_solver}.
Since the error is evaluated at the final time step, the solutions were not smoothed.
For a fair comparison, the orders of the first-order solvers are lowered by one compared to their second-order counterparts, such that their highest modeled derivatives coincide.

\begin{figure*}[t!]
  \centering
  \includegraphics[width=\linewidth]{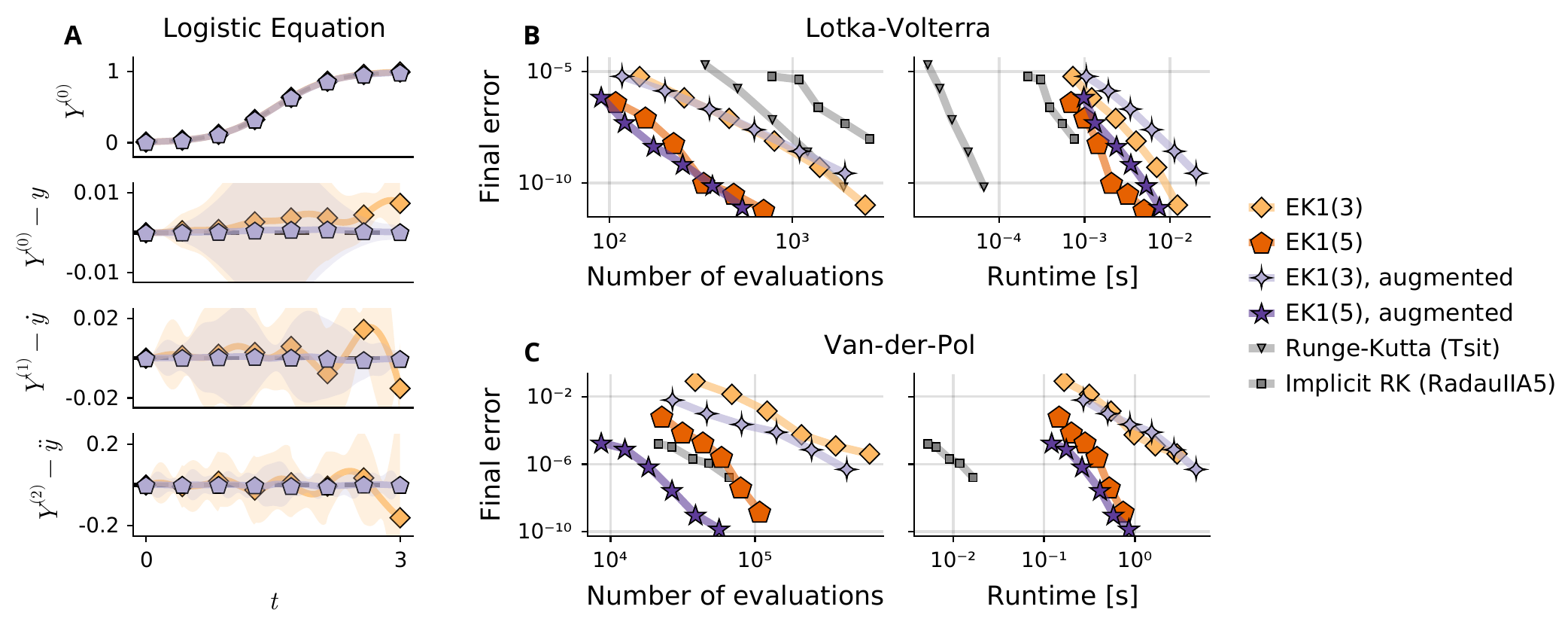}
  \caption{
    \emph{Additional second-derivative information can improve probabilistic solutions.}
    On a fixed time discretization, additional information about second-derivatives reduces the approximation error (A).
    For adaptive-step solvers, it depends on the specific problem.
    On the non-stiff Lotka--Volterra problem, the utility of the additional information seems limited (B).
    However, the benefit of second-derivative information on the stiff Van--der--Pol problem outweighs the additional computational cost and leads to reduced runtimes (C).
  }
  \label{fig:fdb-wps}
\end{figure*}

\paragraph{Results}
The work-precision diagrams in \cref{fig:ode2} show that
second-order ODEs are solved both more efficiently and more accurately than their first-order counterparts.
We observe not only an improvement in absolute runtime, but also a reduced error even for comparable numbers of vector-field evaluations.
\Cref{fig:ode2} also compares the solvers to well-established non-probabilistic methods, including an explicit Runge--Kutta solver
\citep[\texttt{Vern6};][]{verner2010numerically},
a Nyström method
\citep[\texttt{DPRKN6};][]{dormand1987runge},
and an implicit solver
\citep[\texttt{RadauIIA5};][]{radau}.
While they require a comparable number of vector-field evaluations, they exhibit a reduced absolute runtime.
Since probabilistic solvers have the same cubic complexity as the classic, implicit \texttt{RadauIIA5},
we suspect that this discrepancy is partly due to the well-optimized implementation of the \mbox{DifferentialEquations.jl} library \citep{rackauckas2017differentialequations}.
On the other hand, probabilistic ODE solvers provide strictly more functionality than non-probabilistic methods, thus a certain increase in runtime is expected.
As demonstrated by this case study, this paper further reduces the gap between probabilistic and non-probabilistic methods by providing ODE filters for second-order differential equations.

\subsection{First-Order ODEs with Additional Second-Derivative Information}
\label{sec:chainrule}
In this section, we augment the probabilistic solver with additional second-derivative information, that can be derived from a standard, first-order problem.
For this, consider an autonomous, explicit, first-order ODE
\begin{equation}
  \dot{y}(t) = f \left( y(t) \right), \qquad \forall t \in [0, T],
  \label{eq:fdb:1}
\end{equation}
with vector field \(f: \mathbb{R}^d \to \mathbb{R}^d\) and initial value
\(y(0) = y_0 \in \mathbb{R}^d\).
Second derivatives of the true solution can be derived from \cref{eq:fdb:1} by differentiating both sides and applying the chain rule.
We obtain
\begin{equation}
  \ddot{y}(t) = J_f(y(t)) \cdot f \left( y(t) \right),
  \label{eq:fdb:2}
\end{equation}
where \(J_f\) is the Jacobian of \(f\).

\paragraph{Solver Setup}
\Cref{eq:fdb:1,eq:fdb:2} motivate a measurement model
\(z(t, Y) := \left[ z_1(t, Y)\T, z_2(t, Y)\T \right]\T\),
\begin{subequations}
  \begin{align}
    z_1(t, Y) &:= Y^{(1)} - f \left( Y^{(0)} \right), \\
    z_2(t, Y) &:= Y^{(2)} - J_f(Y^{(0)}) f \left( Y^{(0)} \right).
  \end{align}
\end{subequations}
In this evaluation, we consider exact linearizations of both \(z_1\) and \(z_2\) (computed with automatic differentiation).
Furthermore, the solvers update on both measurement models in a single, joint update step.

\paragraph{Fixed-Step Results}
We first evaluate the proposed method in a simplified setting to visualize the effect of additional second-derivative information.
To this end, consider the logistic ODE (defined in \cref{sec:prob:logistic}), and fixed-step solvers with \(\Delta t = 3/7\).
\Cref{fig:fdb-wps} (A) shows the results.
Both solvers approximate the true solution, but the more informed solver achieves lower approximation errors and has reduced uncertainties.

\begin{figure*}[t!]
  \centering
  \includegraphics[width=\textwidth]{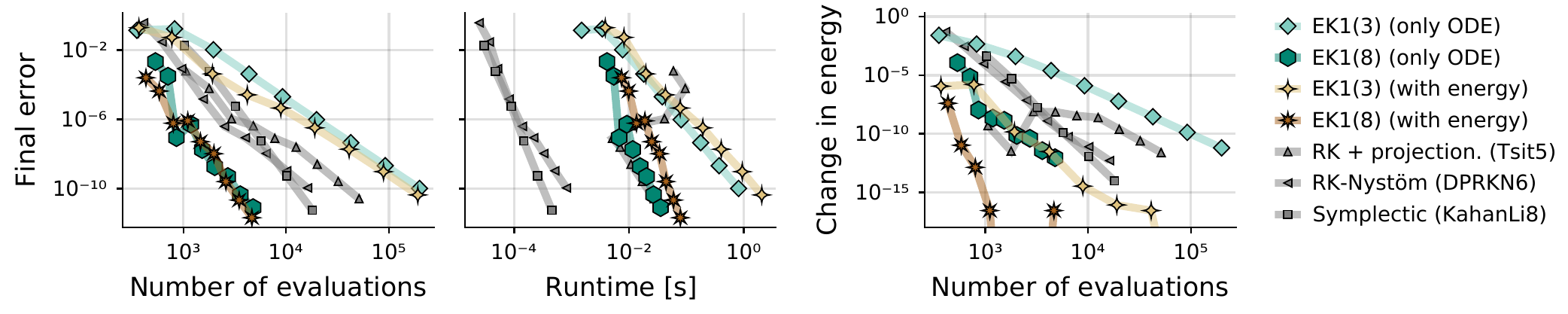}
  \caption{
    \emph{Work-precision diagram of numerical solvers with and without energy conservation.}
    Additional information about the total energy in the dynamical Hénon--Heiles system can improve the accuracy of the solution (left).
    This comes with additional computational cost and increases the runtime (center).
    But as a result, the total energy is conserved more strictly and solutions become more physically meaningful (right).
  }
  \label{fig:henon_heiles_workprecision}
\end{figure*}

\paragraph{Adaptive-Step Results}
Next, we evaluate the proposed method
on the non-stiff Lotka--Volterra problem
(\cref{sec:prob:lotkavolterra})
and the stiff Van--der--Pol model
(\cref{sec:prob:vanderpol}),
in conjunction with adaptive step-size selection and a dynamic diffusion model
\citep{bosch20_calib_adapt_probab_ode_solver}.
\Cref{fig:fdb-wps} shows the resulting work-precision diagrams.
On the Lotka--Volterra problem (B), we observe that additional information does not strictly lead to improvements.
Here, the original \texttt{EK1} solvers seem preferable.
On the other hand, the additional second-derivative information leads to increased accuracy and even to a reduction in the number of vector-field evaluations on the stiff Van--der--Pol problem (C).

\subsection{Systems with Conserved Quantities}
\label{sec:manifold}

In this case study, we demonstrate how additional knowledge about conserved quantities of the modeled dynamical system can be provided to the probabilistic solver.
To this end, we consider \emph{Hamiltonian problems}, a particular class of dynamical systems of the form
\begin{equation}
  \label{eq:hamiltonian-ode}
  \dot{p} = - \frac{\partial H}{\partial q} (p, q), \qquad
  \dot{q} = \frac{\partial H}{\partial p} (p, q),
\end{equation}
where the Hamiltonian
\(H: \mathbb{R}^d \times \mathbb{R}^d \to \mathbb{R}\)
describes the total energy in the dynamical system.
Hamiltonian problems form an important class of ODEs in the context of geometric numerical integration
\citep{hairer2006geometric}
since their
trajectories
\emph{preserve the Hamiltonian}.
That is, for a solution \((p(t), q(t))\) of such problems, the Hamiltonian \(H(p(t), q(t))\) is constant, and it holds
\begin{equation}
  \label{eq:hamiltonian-energy}
  g\left( p(t), q(t) \right) := H(p(t), q(t)) - H(p(0), q(0)) \equiv 0.
\end{equation}
Geometric integrators aim to preserve this structure in their numerical approximation.
In the following, we present a probabilistic solver for Hamiltonian problems that includes this additional information into its inference process to improve its solution estimates.

\paragraph{Solver Setup}
The problems considered in this section can all be written as second-order ODEs, with \((\dot{y}, y) := (p, q)\).
Together with the conservation law of \cref{eq:hamiltonian-energy}, this motivates a partitioned measurement model
\(z(t, Y) := [z_1(t, Y)\T, z_2(t, Y)\T]\T\),
with
\begin{subequations}
  \label{eq:manifold-measurement-setup}
  \begin{align}
    z_1(t, Y) &:= Y^{(2)} - f \left( Y^{(0)} \right), \\
    z_2(t, Y) &:= g(Y^{(1)}, Y^{(0)}),
  \end{align}
\end{subequations}
where \(f\) denotes the vector field of the corresponding ODE.
As in the previous section, all considered methods rely on exact linearizations of the measurement models.
In addition, the solvers perform a \emph{partitioned} EKF update.
That is, they separately linearize and update first on the ODE information \(z_1\) and then on the conserved quantity \(z_2\) -- a procedure that parallels established ``projection methods'' used with non-probabilistic ODE solvers
\citep[Section IV.4]{hairer2006geometric}.

\begin{figure}[t!]
  \centering
  \includegraphics[width=\linewidth]{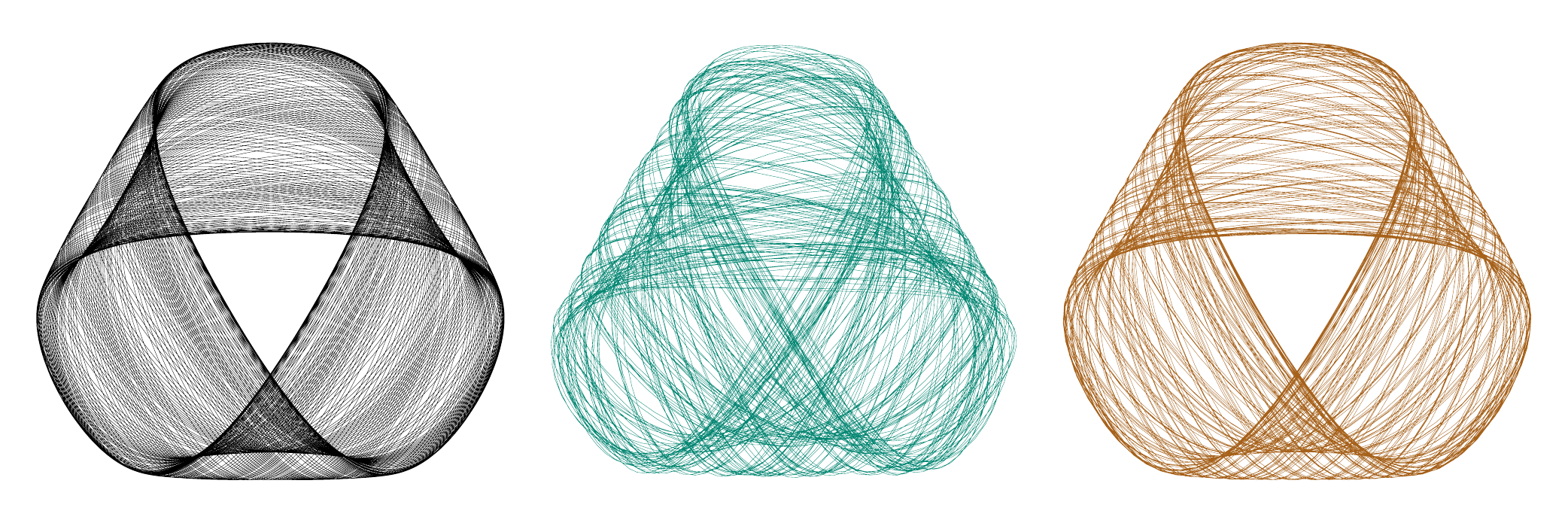}
  \caption{
    \emph{Conservation stabilizes long simulations.}
    Probabilistic numerical simulations of the Hénon--Heiles problem over long time horizons, computed with adaptive steps and low precision,
    deteriorate over time (\textcolor{black}{middle}) and deviate strongly from the true trajectory (left).
    By including energy-preservation into the solver, long-term simulations become more accurate (\textcolor{black}{right}).
  }
  \label{fig:hh_longterm}
\end{figure}

\paragraph{Problem Setting}
We mainly consider the Hénon--Heiles model which describes a star moving around a galactic center \citep{henonheiles}.
The full problem definition is given in \cref{sec:prob:henonheiles}.
We compare probabilistic solvers with and without additional information about the conservation of energy, for various orders (\(q \in \{3,8\}\)).
All solvers use adaptive steps and dynamic diffusion models.
Since we evaluate the error at the final time point, smoothing is not required.

\paragraph{Results}
\Cref{fig:henon_heiles_workprecision} shows the results in multiple work-precision diagrams.
We observe that the additional information leads, in some configurations, to improved accuracies, but comes with an increase in absolute runtime.
However, the probabilistic solvers enforce the conservation of energy very strictly --
even in comparison to non-probabilistic approaches that are particularly well suited for this problem setting, including a Runge--Kutta solver
\citep[\texttt{Tsit5};][]{tsit5}
combined with a projection method
\citep[Section IV.4]{hairer2006geometric},
a Runge--Kutta--Nyström solver
\citep[\texttt{DPRKN6};][]{dormand1987runge},
and a symplectic integrator
\citep[\texttt{KahanLi8};][]{kahan1997composition}.
This structural preservation is of major concern to obtain physically meaningful solutions and stable long-term simulations of Hamiltonian systems \citep{hairer2006geometric}.
The conservation of energy is therefore often of higher importance than a sole reduction in the (Euclidean) error.
Following this motivation, \cref{fig:hh_longterm} shows how energy preservation stabilizes long-term simulations with probabilistic solvers.
%
\begin{figure}[t!]
  \centering
  \includegraphics[width=\linewidth]{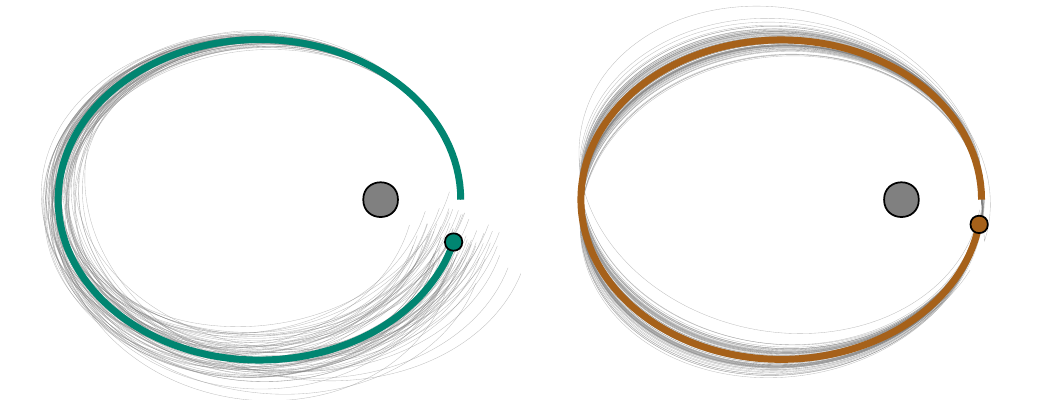}
  \caption{
    \emph{Energy preservation affects the covariances.}
    (For a clearer visualization of the samples, covariances are inflated by a factor of 3 (left) and 300 (right).)
    The samples of a standard probabilistic solution of the Kepler problem appear physically implausible (left).
    By informing the solver about the conservation of energy and angular momentum, the posterior distribution becomes more physically meaningful (right).
  }
  \label{fig:manifold-samples}
\end{figure}
Finally, \cref{fig:manifold-samples} demonstrates on the Kepler problem (defined in \cref{sec:prob:kepler}) how physical information influences not only the mean of the solution estimate, but also its covariances.

\subsection{Differential-Algebraic Equations}
In our final case study, we demonstrate how flexible information operators can be used to extend probabilistic solvers to completely new problem classes.
To this end, we consider systems of the form
\begin{equation} \label{eq:dae}
  M \dot{y}(t) = f \left( y(t) \right), \qquad \forall t \in [0, T],
\end{equation}
with vector field \(f : \mathbb{R}^{d} \to \mathbb{R}^d\), initial values \(y(0) = y_0\),
and \emph{mass matrix} \(M \in \mathbb{R}^{d \times d}\).
If \(M\) is singular, the system can not be rewritten as a regular ODE and we call \cref{eq:dae} a differential-algebraic equation (DAE).
For instance, in the Robertson DAE considered in this case study, we have
\(M = \operatorname{diag}\left( [1, 1, 0] \right)\).
The system thus describes two ODEs and one algebraic equation.
%

DAEs arise naturally in many dynamical systems, such as multi-body dynamics, chemical kinetics, or optimal control
\citep{brenan1996numerical}.
Their numerical simulation is notoriously challenging and often requires specialized methods;
only a specific subset of classic ODE solvers is able to solve the problem given in
\cref{eq:dae}
\citep{daesarenotodes}.
To the best of our knowledge, this work presents the first \emph{probabilistic} DAE solver.

\paragraph{Solver Setup}
To encode the DAE information of \cref{eq:dae}, we define a
measurement model
\begin{equation}
  z(t, Y) := M Y^{(1)} - f \left( Y^{(0)} \right).
\end{equation}
In our experiments, we consider exact linearizations
\begin{equation}
  H := M \cdot E_1 - J_f \left( Y^{(0)} \right) \cdot E_0,
\end{equation}
together with adaptive step-size selection and dynamically calibrated diffusions.
Smoothing is not required.

\label{sec:daes}
\begin{figure}[t!]
  \centering
  \includegraphics[width=\linewidth]{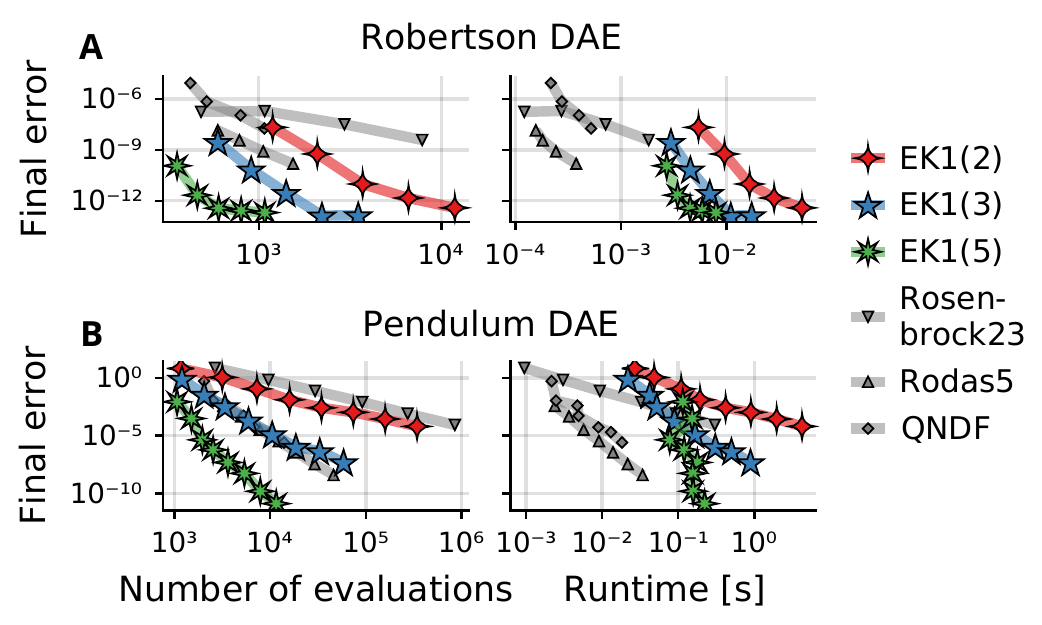}
  \caption{
    \emph{With the correct information operator, probabilistic ODE solvers can solve semi-explicit DAEs.}
  }
  \label{fig:dae}
\end{figure}

\paragraph{Experiment and Results}
To investigate the utility of the proposed methods, we compare probabilistic solvers with various orders (\(q \in \{2,3,5\}\)) to three non-probabilistic DAE solvers:
Rosenbrock methods of order 2 and 5
\citep[\texttt{Rosenbrock23} \& \texttt{Rodas5};][]{hairer1987solving}
and an adaptive-order multistep method
\citep[\texttt{QNDF};][]{shampine1997matlab}.
All methods are evaluated on the stiff Robertson DAE and on a non-stiff pendulum DAE (defined in \cref{sec:prob:rober,sec:prob:pendulum}).
%
\Cref{fig:dae} shows the resulting work-precision diagrams.
As one would expect, increasing the number of steps leads to reduced error.
In addition, we observe higher convergence rates for solvers of higher order.
While the proposed solvers display higher runtimes than their classic counterparts, the differences are comparable to our results in the other case studies and, we suspect, are strongly influenced by implementation details.
In the number of vector-field evaluations, probabilistic and non-probabilistic solvers appear comparable.
In summary, the proposed probabilistic solvers demonstrate good performance on the considered DAEs.


\section{CONCLUSION}
We have shown how to improve ODE solvers by drawing on various sources of information, within the framework of probabilistic numerics.
The proposed algorithm performs efficient inference with extended Kalman filtering and can leverage existing methods for uncertainty calibration and step-size adaptation.
In four case studies, we demonstrated how information about problem structure, additional derivatives, and conserved quantities can be used to improve the solver performance and the quality of the posterior distributions.
By providing a flexible and efficient means to encode mechanistic knowledge beyond the ODE itself, our proposed framework further reduces the gap between probabilistic and non-probabilistic methods and thereby enriches the interface of mechanistic inference and simulation.

\subsubsection*{Acknowledgements}
The authors gratefully acknowledge financial support by the German Federal Ministry of Education and Research (BMBF) through Project ADIMEM (FKZ 01IS18052B), and
financial support by the European Research Council through ERC StG Action 757275 / PANAMA; the DFG Cluster of Excellence “Machine Learning - New Perspectives for Science”, EXC 2064/1, project number 390727645; the German Federal Ministry of Education and Research (BMBF) through the Tübingen AI Center (FKZ: 01IS18039A); and funds from the Ministry of Science, Research and Arts of the State of Baden-Württemberg.
The authors also thank the International Max Planck Research School for Intelligent Systems (IMPRS-IS) for supporting N. Bosch.
The authors are grateful to Nicholas Krämer for many valuable discussions and thank Jonathan Wenger for helpful feedback on the manuscript.

\bibliography{references}

\begin{thebibliography}{}

\bibitem[Abdulle and Garegnani, 2020]{Abdulle2020}
Abdulle, A. and Garegnani, G. (2020).
\newblock Random time step probabilistic methods for uncertainty quantification
  in chaotic and geometric numerical integration.
\newblock {\em Statistics and Computing}, 30.

\bibitem[Bezanson et~al., 2017]{bezanson2017julia}
Bezanson, J., Edelman, A., Karpinski, S., and Shah, V.~B. (2017).
\newblock Julia: A fresh approach to numerical computing.
\newblock {\em SIAM review}, 59.

\bibitem[Bosch et~al., 2021]{bosch20_calib_adapt_probab_ode_solver}
Bosch, N., Hennig, P., and Tronarp, F. (2021).
\newblock Calibrated adaptive probabilistic {ODE} solvers.
\newblock In {\em Proceedings of The 24th International Conference on
  Artificial Intelligence and Statistics}, volume 130 of {\em Proceedings of
  Machine Learning Research}. PMLR.

\bibitem[Brenan et~al., 1996]{brenan1996numerical}
Brenan, K., Campbell, S., Campbell, S., and Petzold, L. (1996).
\newblock {\em Numerical Solution of Initial-Value Problems in
  Differential-Algebraic Equations}.
\newblock Classics in Applied Mathematics. Society for Industrial and Applied
  Mathematics.

\bibitem[Calderhead et~al., 2009]{calderhead09_acc_bayes_inf}
Calderhead, B., Girolami, M., and Lawrence, N. (2009).
\newblock Accelerating {Bayesian} inference over nonlinear differential
  equations with {Gaussian} processes.
\newblock In {\em Advances in Neural Information Processing Systems},
  volume~21. Curran Associates, Inc.

\bibitem[Chkrebtii et~al.,
  2016]{chkrebtii13_bayes_solut_uncer_quant_differ_equat}
Chkrebtii, O.~A., Campbell, D.~A., Calderhead, B., and Girolami, M.~A. (2016).
\newblock {Bayesian Solution Uncertainty Quantification for Differential
  Equations}.
\newblock {\em Bayesian Analysis}, 11(4).

\bibitem[Cockayne et~al., 2019]{cockayne17_bayes_probab_numer_method}
Cockayne, J., Oates, C., Sullivan, T., and Girolami, M. (2019).
\newblock {Bayesian} probabilistic numerical methods.
\newblock {\em SIAM review}, 61.

\bibitem[Conrad et~al., 2017]{Conrad2017}
Conrad, P.~R., Girolami, M., S{\"a}rkk{\"a}, S., Stuart, A., and Zygalakis, K.
  (2017).
\newblock Statistical analysis of differential equations: introducing
  probability measures on numerical solutions.
\newblock {\em Statistics and Computing}, 27.

\bibitem[Dormand and Prince, 1987]{dormand1987runge}
Dormand, J. and Prince, P. (1987).
\newblock {R}unge-{K}utta-{N}ystrom triples.
\newblock {\em Computers \& Mathematics with Applications}, 13.

\bibitem[Gustafsson et~al., 1988]{Gustafsson1988}
Gustafsson, K., Lundh, M., and S{\"o}derlind, G. (1988).
\newblock A {PI} stepsize control for the numerical solution of ordinary
  differential equations.
\newblock {\em BIT Numerical Mathematics}, 28.

\bibitem[Hairer et~al., 2006]{hairer2006geometric}
Hairer, E., Lubich, C., and Wanner, G. (2006).
\newblock {\em Geometric Numerical Integration: Structure-Preserving Algorithms
  for Ordinary Differential Equations}.
\newblock Springer Series in Computational Mathematics. Springer Berlin
  Heidelberg.

\bibitem[Hairer et~al., 1993]{hairer2008solving}
Hairer, E., Norsett, S., and Wanner, G. (1993).
\newblock {\em Solving Ordinary Differential Equations I: Nonstiff Problems},
  volume~8.
\newblock Springer-Verlag.

\bibitem[Hairer and Wanner, 1996]{hairer1987solving}
Hairer, E. and Wanner, G. (1996).
\newblock {\em Solving Ordinary Differential Equations II. Stiff and
  Differential-Algebraic Problems}, volume~14.
\newblock Springer-Verlag.

\bibitem[Hairer and Wanner, 1999]{radau}
Hairer, E. and Wanner, G. (1999).
\newblock Stiff differential equations solved by {R}adau methods.
\newblock {\em Journal of Computational and Applied Mathematics}, 111.

\bibitem[Hennig et~al., 2015]{hennig15_probab_numer_uncer_comput}
Hennig, P., Osborne, M.~A., and Girolami, M. (2015).
\newblock Probabilistic numerics and uncertainty in computations.
\newblock {\em Proceedings. Mathematical, physical, and engineering sciences},
  471.

\bibitem[{Henon} and {Heiles}, 1964]{henonheiles}
{Henon}, M. and {Heiles}, C. (1964).
\newblock {The applicability of the third integral of motion: Some numerical
  experiments}.
\newblock {\em The Astronomical Journal}, 69.

\bibitem[Julier and Uhlmann, 2004]{1271397}
Julier, S. and Uhlmann, J. (2004).
\newblock Unscented filtering and nonlinear estimation.
\newblock {\em Proceedings of the IEEE}, 92.

\bibitem[Kahan and Li, 1997]{kahan1997composition}
Kahan, W. and Li, R.-C. (1997).
\newblock Composition constants for raising the orders of unconventional
  schemes for ordinary differential equations.
\newblock {\em Mathematics of computation}, 66.

\bibitem[Kersting et~al., 2020]{kersting18_conver_rates_gauss_ode_filter}
Kersting, H., Sullivan, T.~J., and Hennig, P. (2020).
\newblock Convergence rates of {Gaussian} {ODE} filters.
\newblock {\em Statistics and Computing}, 30.

\bibitem[Kr{\"a}mer and Hennig, 2020]{kraemer20_stabl_implem_probab_ode_solver}
Kr{\"a}mer, N. and Hennig, P. (2020).
\newblock Stable implementation of probabilistic {ODE} solvers.
\newblock {\em arXiv:2012.10106 [stat.ML]}.

\bibitem[Nystr{\"o}m, 1925]{nystrom1925numerische}
Nystr{\"o}m, E. (1925).
\newblock {\em {\"U}ber die numerische integration von
  differentialgleichungen}.
\newblock Acta Societatis Scientiarum Fennicae. Finn. Literaturges.

\bibitem[Oates and Sullivan, 2019]{DBLP:journals/sac/OatesS19}
Oates, C.~J. and Sullivan, T.~J. (2019).
\newblock A modern retrospective on probabilistic numerics.
\newblock {\em Statistics and Computing}, 29.

\bibitem[Petzold, 1982]{daesarenotodes}
Petzold, L. (1982).
\newblock Differential/algebraic equations are not {ODE}'s.
\newblock {\em Siam Journal on Scientific and Statistical Computing}, 3.

\bibitem[Rackauckas and Nie, 2017]{rackauckas2017differentialequations}
Rackauckas, C. and Nie, Q. (2017).
\newblock {DifferentialEquations.jl} -- a performant and feature-rich ecosystem
  for solving differential equations in {J}ulia.
\newblock {\em Journal of Open Research Software}, 5.

\bibitem[Raitoharju and Piché, 2019]{8861457}
Raitoharju, M. and Piché, R. (2019).
\newblock On computational complexity reduction methods for {K}alman filter
  extensions.
\newblock {\em IEEE Aerospace and Electronic Systems Magazine}, 34.

\bibitem[Raitoharju et~al., 2016]{partitionedupdateKF}
Raitoharju, M., Piché, R., Ala-Luhtala, J., and Ali-Löytty, S. (2016).
\newblock Partitioned update {K}alman filter.
\newblock {\em Journal of Advances in Information Fusion}, 11.

\bibitem[Raitoharju et~al., 2017]{RAITOHARJU2017289}
Raitoharju, M., Ángel F.~García-Fernández, and Piché, R. (2017).
\newblock {K}ullback--{L}eibler divergence approach to partitioned update
  {K}alman filter.
\newblock {\em Signal Processing}, 130.

\bibitem[S{\"{a}}rkk{\"{a}}, 2013]{sarkka_bayesianfilteringandsmoothing}
S{\"{a}}rkk{\"{a}}, S. (2013).
\newblock {\em Bayesian Filtering and Smoothing}, volume~3 of {\em Institute of
  Mathematical Statistics textbooks}.
\newblock Cambridge University Press.

\bibitem[Schober et~al.,
  2019]{schober16_probab_model_numer_solut_initial_value_probl}
Schober, M., S{\"a}rkk{\"a}, S., and Hennig, P. (2019).
\newblock A probabilistic model for the numerical solution of initial value
  problems.
\newblock {\em Statistics and Computing}, 29.

\bibitem[Shampine and Reichelt, 1997]{shampine1997matlab}
Shampine, L.~F. and Reichelt, M.~W. (1997).
\newblock The {Matlab} {ODE} suite.
\newblock {\em SIAM journal on scientific computing}, 18.

\bibitem[Särkkä and Solin, 2019]{sarkka_solin_2019}
Särkkä, S. and Solin, A. (2019).
\newblock {\em Applied Stochastic Differential Equations}.
\newblock Institute of Mathematical Statistics Textbooks. Cambridge University
  Press.

\bibitem[Teymur et~al., 2018]{NEURIPS2018_228b2558}
Teymur, O., Lie, H.~C., Sullivan, T., and Calderhead, B. (2018).
\newblock Implicit probabilistic integrators for {ODEs}.
\newblock In {\em Advances in Neural Information Processing Systems},
  volume~31. Curran Associates, Inc.

\bibitem[Tronarp et~al., 2019]{tronarp18_probab_solut_to_ordin_differ}
Tronarp, F., Kersting, H., S{\"{a}}rkk{\"{a}}, S., and Hennig, P. (2019).
\newblock Probabilistic solutions to ordinary differential equations as
  nonlinear {Bayesian} filtering: a new perspective.
\newblock {\em Statistics and Computing}, 29.

\bibitem[Tronarp et~al., 2021]{tronarp20_bayes_ode_solver}
Tronarp, F., S{\"a}rkk{\"a}, S., and Hennig, P. (2021).
\newblock Bayesian {ODE} solvers: the maximum a posteriori estimate.
\newblock {\em Statistics and Computing}, 31.

\bibitem[Tsitouras, 2011]{tsit5}
Tsitouras, C. (2011).
\newblock Runge–{K}utta pairs of order 5 (4) satisfying only the first column
  simplifying assumption.
\newblock {\em Computers \& Mathematics with Applications}, 62.

\bibitem[van~der Pol, 1926]{vanderpol}
van~der Pol, B. (1926).
\newblock On "relaxation-oscillations".
\newblock {\em The London, Edinburgh, and Dublin Philosophical Magazine and
  Journal of Science}, 2.

\bibitem[Verner, 2010]{verner2010numerically}
Verner, J.~H. (2010).
\newblock Numerically optimal {R}unge–{K}utta pairs with interpolants.
\newblock {\em Numerical Algorithms}, 53.

\bibitem[Wenk et~al., 2020]{wenk19_odin}
Wenk, P., Abbati, G., Osborne, M.~A., Sch{\"o}lkopf, B., Krause, A., and Bauer,
  S. (2020).
\newblock {ODIN}: {ODE}-informed regression for parameter and state inference
  in time-continuous dynamical systems.
\newblock In {\em Proceedings of the 34th Conference on Artificial Intelligence
  (AAAI)}, volume~34. AAAI Press.

\end{thebibliography}


\begin{thebibliography}{}

\bibitem[Hairer et~al., 2006]{hairer2006geometric}
Hairer, E., Lubich, C., and Wanner, G. (2006).
\newblock {\em Geometric Numerical Integration: Structure-Preserving Algorithms
  for Ordinary Differential Equations}.
\newblock Springer Series in Computational Mathematics. Springer Berlin
  Heidelberg.

\bibitem[Hairer et~al., 1993]{hairer2008solving}
Hairer, E., Norsett, S., and Wanner, G. (1993).
\newblock {\em Solving Ordinary Differential Equations I: Nonstiff Problems},
  volume~8.
\newblock Springer-Verlag.

\bibitem[Hairer and Wanner, 1996]{hairer1987solving}
Hairer, E. and Wanner, G. (1996).
\newblock {\em Solving Ordinary Differential Equations II. Stiff and
  Differential-Algebraic Problems}, volume~14.
\newblock Springer-Verlag.

\bibitem[{Henon} and {Heiles}, 1964]{henonheiles}
{Henon}, M. and {Heiles}, C. (1964).
\newblock {The applicability of the third integral of motion: Some numerical
  experiments}.
\newblock {\em The Astronomical Journal}, 69.

\bibitem[van~der Pol, 1926]{vanderpol}
van~der Pol, B. (1926).
\newblock On "relaxation-oscillations".
\newblock {\em The London, Edinburgh, and Dublin Philosophical Magazine and
  Journal of Science}, 2.

\end{thebibliography}

\onecolumn
\aistatstitle{Supplementary Material: Pick-and-Mix Information Operators for Probabilistic ODE Solvers}

\appendix

\section{PROBLEM DEFINITIONS}

\subsection{Pleiades}
\label{sec:prob:pleiades}
The Pleiades system describes the motion of seven stars in a plane,
with coordinates \((x_i, y_i)\) and masses \(m_i = i\), \(i=1, \dots, 7\)
\citep[II.10]{hairer2008solving}.
It is given by a second-order ODE
\begin{equation}
  \ddot{x}_i = \sum_{j \neq i} m_j (x_j - x_i) / r_{ij}, \qquad
  \ddot{y}_i = \sum_{j \neq i} m_j (y_j - y_i) / r_{ij}, \qquad
\end{equation}
where \(r_{ij} = \left( (x_i - x_j)^2 + (y_i - y_j)^2 \right)^{3/2}\), for \(i,j=1,\dots,7\),
on the time span \(t \in [0, 3]\),
with initial locations
\begin{subequations}
\begin{align}
  x(0) &= [3,3,-1,-3,2,-2,2], \\
  y(0) &= [3,-3,2,0,0,-4,4],
\intertext{and initial velocities}
\dot{x}(0) &= [0,0,0,0,0,1.75,-1.5], \\
\dot{y}(0) &= [0,0,0,-1.25,1,0,0].
\end{align}
\end{subequations}

\subsection{Logistic Equation}
\label{sec:prob:logistic}
The logistic equation is a simple IVP problem, given as
\begin{equation}
  \dot{y}(t) = 3 y(t) (1-y(t)), \qquad t \in [0,3], \qquad y(0) = 100,
\end{equation}
for which the analytical solution is known to be
\begin{equation}
  y(t) = \frac{\exp(3 t)}{100 - 1 + \exp(3t)}.
\end{equation}

\subsection{Lotka--Volterra}
\label{sec:prob:lotkavolterra}
The Lotka--Volterra model describes the dynamics of biological systems in which two species interact, one as a predator and the other as prey.
The IVP is given by the ODE
\begin{equation}
  \dot{x} = 1.5 x - x y, \qquad \dot{y} = x y -3 y.
\end{equation}
In our experiments, we consider initial values
\(x(0) = 1\), \(y(0) = 1\) and a
time span
\(t \in [0, 7]\).

\subsection{Van--der--Pol}
\label{sec:prob:vanderpol}
The Van der Pol model \citep{vanderpol} describes a non-conservative oscillator with non-linear damping.
In our experiment, we consider a notoriously stiff version of the model, given as
\begin{subequations}%
  \begin{align}%
    \dot{y_1}(t) = y_2(t), \qquad
    \dot{y_2}(t) = 10^6 \left( \left( 1-y_1^2(t) \right) y_2(t) - y_1(t) \right),%
  \end{align}%
\end{subequations}%
on the time span \(t \in [0,10]\), with initial value \(y(0) = [0, \sqrt{3}]\T\).

\subsection{H\'{e}non--Heiles}
\label{sec:prob:henonheiles}
The H\'{e}non-Heiles model describes a star moving around a galactic center, with its motion restricted to a plane
\citep{henonheiles}.
It is defined by a Hamiltonian
\begin{equation}
  H(p, q) =
  \left[ \frac{1}{2} \left( p_1^2 + p_2^2 \right) \right] +
  \left[ \frac{1}{2} \left( q_1^2 + q_2^2 \right) + q_1^2 q_2 - \frac{1}{3} q_2^3 \right],
\end{equation}
which describes the kinetic and potential energy of the star with
velocity \(p\) and location \(q\).
With \(y(t) := q(t)\), we write the Hénon-Heiles problem as an IVP with second-order ODE, as
\begin{subequations}
\begin{align}
  \ddot{y}_1(t) &= -y_1(t) - 2 y_1(t) y_2(t), \\
  \ddot{y}_2(t) &= y_2^2(t) - y_2(t) - y_1^2(t),
\end{align}
\end{subequations}
on the time span \(t \in [0, 1000]\), with initial values
\(y(0) = (0, 0.1)\), \(\dot{y}(0) = (0.5, 0)\).
It further holds
\begin{equation}
  g(\dot{y}(t), y(t)) := H(\dot{y}(t), y(t)) - H(\dot{y}_0(t), y_0(t)) = 0,
\end{equation}
by conservation of the Hamiltonian \citep{hairer2006geometric}.

\subsection{Kepler Problem}
\label{sec:prob:kepler}
The Kepler problem is a special case of the two-body problem in celestial
mechanics, and can be used to describe the movement of a planet around a star.
It is given by a Hamiltonian \(H : \mathbb{R}^2 \times \mathbb{R}^2 \to \mathbb{R}\), with
\begin{equation}
  H \left( p(t), q(t) \right) = \frac{\|p(t)\|^2}{2} - \frac{1}{\|q(t)\|}.
\end{equation}
With \(y(t) := q(t)\) and \(\dot{y}(t) := p(t)\), it induces the second-order ODE
\begin{equation}
  \ddot{y}(t) = - \frac{y(t)}{\|y(t)\|^3}.
\end{equation}
In our experiments, the Kepler problem is solved on the time span
\(t \in \left[ 0, \frac{99}{100} \cdot 2 \pi \right] \),
with initial values
\(y(0) = [0.4, 0]\), \(\dot{y}(0) = [0, 2]\).
In addition to conserving the Hamiltonian, the Kepler system conserves angular momentum:
\begin{equation}
  L \left( p(t), q(t) \right) = q_1(t)p_2(t) - q_2(t)p_1(t).
\end{equation}
Thus, it holds
\begin{equation}
  g(\dot{y}(t), y(t)) :=
  \begin{bmatrix}
  H(\dot{y}(t), y(t)) - H(\dot{y}(0), y(0)) \\
  L(\dot{y}(t), y(t)) - L(\dot{y}(0), y(0))
  \end{bmatrix} = 0.
\end{equation}

\subsection{Robertson DAE}
\label{sec:prob:rober}
The Robertson DAE describes a system of chemical reactions and is a very popular problem to evaluate stiff ODE and DAE solvers \citep{hairer1987solving}.
As a DAE, it is given by the equations
\begin{subequations}
  \begin{align}
    y_1(t) &= -0.04 y_1(t) + 10^4 y_2(t) y_3(t), \\
    y_2(t) &= 0.04 y_1(t) + 10^4 y_2(t) y_3(t) - (3 \cdot 10^7) y_2(t)^2, \\
    0 &= y_1(t) + y_2(t) + y_3(t) - 1,
  \end{align}
\end{subequations}
and it therefore has a singular mass matrix of the form
\[M = \begin{bmatrix} 1 & 0 & 0 \\ 0 & 1 & 0 \\ 0 & 0 & 0 \end{bmatrix}.\]
We consider an initial value
\(y(0) = [1,0,0]\),
and while the system is most often simulated on the time span
\(t \in [0, 10^5]\),
we solve it on \(t \in [0, 10^2]\) since we found the final error to be more informative in this setting since the values did then not saturate yet.

\subsection{Pendulum DAE}
\label{sec:prob:pendulum}
A pendulum can be described in Cartesian coordinates with the following, index-reduced DAE \citep{hairer1987solving}
\begin{subequations}
  \begin{align}
    \dot{x}(t) &= v_x,\\
    \dot{v_x}(t) &= x T,\\
    \dot{y}(t) &= v_y,\\
    \dot{v_y}(t) &= y T - g,\\
    0 &= 2 ( v_x^2 + v_y^2 + y (yT - g) + T x^2).
  \end{align}
\end{subequations}
In our experiments, we consider initial values
\(x(0) = 1\),
\(v_x(0) = 0\),
\(y(0) = 0\),
\(v_y(0) = 0\),
\(T(0) = 0\),
and the gravitational acceleration
\(g = 9.81\).
We simulate the system on the time span \(t \in [0, 10]\).

\end{document}